\documentclass[11pt,a4paper]{article}
\usepackage[hyperref]{acl2021}
\usepackage{times}
\usepackage{latexsym}
\usepackage[printwatermark]{xwatermark}
\usepackage{amsmath,amssymb}
\usepackage{amsfonts}
\usepackage{booktabs}
\usepackage{multirow,multicol}
\usepackage{times}
\usepackage{pifont}
\usepackage{xcolor}
\usepackage{xspace}
\usepackage{csquotes}

\usepackage{textcomp}
\usepackage{fontawesome}
\usepackage{stfloats}
\usepackage{tabularx}
\usepackage{fontawesome}
\usepackage{scalerel}

\newcommand{\noi}{\textsc{nOI}\xspace}
\newcommand{\oni}{\textsc{OnI}\xspace}
\newcommand{\oi}{\textsc{OI}\xspace}
\newcommand{\toxt}{\textsc{ToxTrig}\xspace}
\newcommand{\InvRat}{\textsc{InvRat }}
\newcommand{\InvRa}{\textsc{InvRat}}

\newcommand{\aae}{\textsc{AAE}\xspace}
\newcommand{\wae}{\textsc{wae}\xspace}

\newcommand{\lmixin}{\textsc{LMixin}\xspace}

\DeclareFontFamily{U}{skulls}{}
\DeclareFontShape{U}{skulls}{m}{n}{ <-> skull }{}

\usepackage{microtype}

\aclfinalcopy 

\title{Mitigating Biases in Toxic Language Detection through \\ Invariant Rationalization}

\author{
Yung-Sung Chuang$^{1,2}$ $    $ Mingye Gao$^2$ $    $
Hongyin Luo$^2$ $    $ \\ \textbf{James Glass}$^2$ $    $ \textbf{Hung-yi Lee}$^1$  $    $ \textbf{Yun-Nung Chen}$^1$  $    $  \textbf{Shang-Wen Li}$^3$\thanks{* Work is not related to employment at Amazon.} \\
  $^{1}$National Taiwan University, $^{2}$MIT CSAIL, $^{3}$Amazon AI  \\
  {\tt \{yungsung, mingye, hyluo, glass\}@mit.edu,} \\
  {\tt hungyilee@ntu.edu.tw, y.v.chen@ieee.org, shangwel@amazon.com}\\
  }

\date{}

\begin{document}
\maketitle
\begin{abstract}
Automatic detection of toxic language plays an essential role in protecting social media users, especially minority groups, from verbal abuse. However, biases toward some attributes, including gender, race, and dialect, exist in most training datasets for toxicity detection. The biases make the learned models unfair and can even exacerbate the marginalization of people. Considering that current debiasing methods for general natural language understanding tasks cannot effectively mitigate the biases in the toxicity detectors, we propose to use invariant rationalization (\InvRa), a game-theoretic framework consisting of a rationale generator and predictors, to rule out the spurious correlation of certain syntactic patterns (e.g., identity mentions, dialect) to toxicity labels. We empirically show that our method yields lower false positive rate in both lexical and dialectal attributes than previous debiasing methods.\footnote{The source code is available at \url{https://github.com/voidism/invrat_debias}.}
\end{abstract}

\section{Introduction}
As social media becomes more and more popular in recent years, many users, especially the minority groups, suffer from verbal abuse and assault. To protect these users from online harassment, it is necessary to develop a tool that can automatically detect the toxic language in social media. In fact, many toxic language detection (TLD) systems have been proposed in these years based on different models, such as support vector machines (SVM) \citep{gaydhani2018detecting}, bi-directional long short-term memory (BiLSTM) \citep{bojkovsky2019stufiit}, logistic regression \citep{davidson2017automated} and fine-tuning BERT \citep{d2020bert}.

However, the existing TLD systems exhibit some problematic and discriminatory behaviors \citep{zhou2021challenges}. Experiments show that the tweets containing certain surface markers, such as identity terms and expressions in African American English (AAE), are more likely to be classified as hate speech by the current TLD systems \citep{davidson2017automated, xia2020demoting}, although some of them are not actually hateful. Such an issue is predominantly attributed to the biases in training datasets for the TLD models; when the models are trained on the biased datasets, these biases are inherited by the models and further exacerbated during the learning process \citep{zhou2021challenges}. The biases in TLD systems can make the opinions from the members of minority groups more likely to be removed by the online platform, which may significantly hinder their experience as well as exacerbate the discrimination against them in real life.  

So far, many debiasing methods have been developed to mitigate biases in learned models, such as data re-balancing \citep{dixon2018measuring}, residual fitting \citep{he2019unlearn, clark2019don}, adversarial training \citep{xia2020demoting} and data filtering approach \citep{bras2020adversarial, zhou2021challenges}. While most of these works are successful on other natural language processing (NLP) tasks, their performance on debasing the TLD tasks are unsatisfactory \citep{zhou2021challenges}. A possible reason is that the toxicity of language is more subjective and nuanced than general NLP tasks that often have unequivocally correct labels \citep{zhou2021challenges}. As current debiasing techniques reduce the biased behaviors of models by correcting the training data or measuring the difficulty of modeling them, which prevents models from capturing spurious and non-linguistic correlation between input texts and labels, the nuance of toxicity annotation can make such techniques insufficient for the TLD task.  

In this paper, we address the challenge by combining the TLD classifier with the selective rationalization method, which is widely used to interpret the predictions of complex neural networks. Specifically, we use the framework of Invariant Rationalization (\InvRa) \citep{chang2020invariant} to rule out the syntactic and semantic patterns in input texts that are highly but spuriously correlated with the toxicity label, and mask such parts during inference. Experimental results show that \InvRat successfully reduce the lexical and dialectal biases in the TLD model with little compromise on overall performance. Our method avoids superficial correlation at the level of syntax and semantics, and makes the toxicity detector learn to use generalizable features for prediction, thus effectively reducing the impact of dataset biases and yielding a fair TLD model. 

\section{Previous works}

\paragraph{Debiasing the TLD Task} Researchers have proposed a range of debiasing methods for the TLD task. Some of them try to mitigate the biases by processing the training dataset. For example, \citet{dixon2018measuring} add additional non-toxic examples containing the identity terms highly correlated to toxicity to balance their distribution in the training dataset. \citet{park2018reducing} use the combination of debiased \textit{word2vec} and gender swap data augmentation to reduce the gender bias in TLD task. \citet{badjatiya2019stereotypical} apply the strategy of replacing the bias sensitive words (BSW) in training data based on multiple knowledge generalization.

Some researchers pay more attention to modifying the models and learning less biased features. \citet{xia2020demoting} use adversarial training to reduce the tendency of the TLD system to misclassify the AAE texts as toxic speech. \citet{mozafari2020hate} propose a novel re-weighting mechanism to alleviate the racial bias in English tweets. \citet{vaidya2020empirical} implement a multi-task learning framework with an attention layer to prevent the model from picking up the spurious correlation between the certain trigger-words and toxicity labels. 

\paragraph{Debiasing Other NLP Task} There are many methods proposed to mitigate the biases in NLP tasks other than TLD. \citet{clark2019don} train a robust classifier in an ensemble with a bias-only model to learn the more generalizable patterns in training dataset, which are difficult to be learned by the naive bias-only model. \citet{bras2020adversarial} develop AFLITE, an iterative greedy algorithm that can adversarially filter the biases from the training dataset, as well as the framework to support it. \citet{utama2020mind} introduce a novel approach of regularizing the confidence of models on the biased examples, which successfully makes the models perform well on both in-distribution and out-of-distribution data.      

\section{Invariant Rationalization}

\subsection{Basic Formulation for Rationalization} \label{3.1}

We propose TLD debiasing based on \InvRat in this paper. The goal of rationalization is to find a subset of inputs that 1) suffices to yield the same outcome 2) is human interpretable. Normally, we would prefer to find rationale in unsupervised ways because the lack of such annotations in the data. A typical formulation to find rationale is as following: Given the input-output pairs $(\boldsymbol{X}, Y)$ from a text classification dataset, we use a classifier $f$ to predict the labels $f(\boldsymbol{X})$. To extract the rationale here, an intermediate rationale generator $g$ is introduced to find a rationale $\boldsymbol{Z} = g(\boldsymbol{X})$, a masked version of $X$ that can be used to predict the output Y, i.e. maximize mutual information between $\boldsymbol{Z}$ and $Y$.\footnote{Real examples of $\boldsymbol{X}, \boldsymbol{Z}$ can be found in Table~\ref{tab:qual}.} 
\vspace{-5pt}
\begin{equation}
\max _{\boldsymbol{m} \in \mathcal{S}} I(Y ; \boldsymbol{Z}) \quad \text { s.t. } \boldsymbol{Z}=\boldsymbol{m} \odot \boldsymbol{X}
\end{equation}
Regularization loss $\mathcal{L}_{\text{reg}}$ is often applied to keep the rationale sparse and contiguous:
\vspace{-5pt}
\begin{align} \small
    \mathcal{L}_{\text{reg}}=\lambda_1 \left|\frac{1}{N} \mathbb{E}\left[\|\boldsymbol{m}\|_{1}\right]-\alpha\right| + \lambda_2 \mathbb{E}\left[\sum_{n=2}^{N}\left|\boldsymbol{m}_{n}-\boldsymbol{m}_{n-1}\right|\right]
    \label{eq:sparsity}
\end{align}

\vspace{-20pt}
\subsection{The \InvRat Framework} \label{3.2}
\InvRat \citep{chang2020invariant} introduces the idea of \emph{environment} to rationalization. We assume that the data are collected from different environments with different prior distributions. Among these environments, the predictive power of spurious correlated features will be variant, while the genuine causal explanations always have invariant predictive power to $Y$. Thus, the desired rationale should satisfy the following invariant constraint:
\vspace{-5pt}
\begin{equation}
H(Y|\boldsymbol{Z}, E) = H(Y|\boldsymbol{Z}),
\end{equation}
where $E$ is the given environment and $H$ is the cross-entropy between the prediction and the ground truth $Y$. We can use a three-player framework to find the solution for the above equation: an environment-agnostic predictor $f_{i}(\boldsymbol{Z})$, an environment-aware predictor $f_{e}(\boldsymbol{Z}, E)$, and a rationale generator $g(\boldsymbol{X})$. The learning objective of the two predictors are:
\vspace{-5pt}
\begin{align}
    \mathcal{L}_{i}^{*}&=\min _{f_{i}(\cdot)} \mathbb{E}\left[\mathcal{L}\left(Y ; f_{i}(\boldsymbol{Z})\right)\right] \\
    \mathcal{L}_{e}^{*}&=\min _{f_{e}(\cdot,\cdot)} \mathbb{E}\left[\mathcal{L}\left(Y ; f_{e}(\boldsymbol{Z}, E)\right)\right]
\end{align}
In addition to minimizing the invariant prediction loss $\mathcal{L}_{i}^{*}$ and the regularization loss $\mathcal{L}_{\text{reg}}$, the other objective of the rationale generator is to minimize the gap between $\mathcal{L}_{i}^{*}$ and $\mathcal{L}_{e}^{*}$, that is:
\vspace{-5pt}
\begin{equation}
    \min _{g(\cdot)} \mathcal{L}_{i}^{*} + \mathcal{L}_{\text{reg}} + \lambda_{\text{diff}} \cdot \text{ReLU}\left(\mathcal{L}_{i}^{*}-\mathcal{L}_{e}^{*}\right),
    \label{eq:all}
\end{equation}
where ReLU is applied to prevent the penalty when $\mathcal{L}_{i}^{*}$ has been lower than $\mathcal{L}_{e}^{*}$.

\section{\InvRat for TLD Debiasing}

\subsection{TLD Dataset and its Biases}
We apply \InvRat to debiasing TLD task. For clarity, we seed our following description with a specific TLD dataset where we conducted experiment on, hate speech in Twitter created by \citet{founta2018large} and modified by \citet{zhou2021challenges}, and we will show how to generalize our approach. The dataset contains 32K toxic and 54K non-toxic tweets. Following works done by \citet{zhou2021challenges}, we focus on two types of biases in the dataset: lexical biases and dialectal biases. Lexical biases contain the spurious correlation of toxic language with attributes including Non-offensive minority identity (\noi), Offensive minority identity (\oi), and Offensive non-identity (\oni); dialectal biases are relating African-American English (\aae) attribute directly to toxicity. All these attributes are tagged at the document level. We provide more details for the four attributes (\noi, \oi, \oni, and \aae) in Appendix~\ref{app:attributes}.

\subsection{Use \InvRat for Debiasing}
We directly use the lexical and dialectal attributes as the environments in \InvRat for debiasing TLD\footnote{To generalize our method for any other attributes or datasets, one can simply map environments to the attributes in consideration for debiasing.}. 
Under these different environments, the predictive power of spurious correlation between original input texts $\boldsymbol{X}$ and output labels $\boldsymbol{Y}$ will change. Thus, in \InvRa, the rationale generator will learn to exclude the biased phrases that are spurious correlated to toxicity labels from the rationale $\boldsymbol{Z}$. On the other hand, the predictive power for the genuine linguistic clues will be generalizable across environments, so the rationale generator attempts to keep them in the rationale $\boldsymbol{Z}$.

Since there is no human labeling for the attributes in the original dataset, we infer the labels following \citet{zhou2021challenges}. We match $\boldsymbol{X}$ with \textsc{ToxTrig}, a handcrafted word bank collected for \noi, \oi, and \oni; for dialectal biases, we use the topic model from \citet{blodgett2016demographic} to classify $\boldsymbol{X}$ into four dialects: \aae, white-aligned English (\wae), Hispanic, and other.

We build two debiasing variants with the obtained attribute labels, \InvRat (lexical) and \InvRat (dialect). The former is learned with the compound loss function in Equation (\ref{eq:all}) and four lexical-related environment subsets (\noi, \oi, \oni, and none of the above); we train the latter using the same loss function but along with four dialectal environments (\aae, \wae, Hispanic, and other). In both variants, the learned $f_{i}(\boldsymbol{Z})$ is our environment-agnostic TLD predictor that classifies toxic languages based on generalizable clues. Also, in the \InvRat framework, the environment-aware predictor $f_{e}(\boldsymbol{Z}, E)$ needs to access the environment information. We use an additional embedding layer $\text{Emb}_{\text{env}}$ to embed the environment id $e$ into a $n$-dimensional vector $\text{Emb}_{\text{env}}(e)$, where $n$ is the input dimension of the pretrained language model. Word embeddings and $\text{Emb}_{\text{env}}(e)$ are summed to construct the input representation for $f_{e}$.

\section{Experiment}
\subsection{Experiment Settings}
\label{sec:exp_settings}

We leverage RoBERTa-base~\citep{liu2019roberta} as the backbone of our TLD models in experiments. 
$F_1$ scores and false positive rate (FPR) when specific attributes exist in texts are used to quantify TLD and debiasing performance, respectively. The positive label is "toxic" and the negative label is "non-toxic" for computing $F_1$ scores. When evaluating models debiased by \InvRa, we use the following strategy to balance $F_1$ and FPR, and have a stable performance measurement. We first select all checkpoints with $F_1$ scores no less than the best TLD performance in dev set by 3\%. Then, we pick the checkpoint with the lowest dev set FPR among these selected ones to evaluate on the test set. We describe more training details and used hyperparameters in Appendix~\ref{app:details}.

\subsection{Quantitative Debiasing Results}

\begin{table*}[t]
\centering
\small
\begin{tabular}{clc|crcccc|cc}
\toprule
& & Test & \multicolumn{2}{c}{\noi} & \multicolumn{2}{c}{\oi} & \multicolumn{2}{c}{\oni} & \multicolumn{2}{|c}{AAE}\\
\cmidrule(lr){3-3} \cmidrule(lr){4-5} \cmidrule(lr){6-7} \cmidrule(lr){8-9} \cmidrule(lr){10-11}
& & $F_1\uparrow$ & $F_1\uparrow$ & FPR $\downarrow$ & $F_1\uparrow$ & FPR $\downarrow$ & $F_1 \uparrow$ & FPR $\downarrow$ & $F_1 \uparrow$ & FPR $\downarrow$ \\
\midrule
& Vanilla &  92.3$_{0.0}$ & 89.8$_{0.3}$ & 10.2$_{1.3}$ & 98.8$_{0.1}$ & 85.7$_{0.0}$ & 97.3$_{0.1}$ & 64.7$_{0.8}$ & 92.3$_{0.0}$ & 16.8$_{0.3}$\\
& \lmixin-\oni & 85.6$_{2.5}$ & 87.0$_{1.1}$ & 14.0$_{1.5}$ & 98.9$_{0.0}$ & 85.7$_{0.0}$ & 87.9$_{4.5}$ & \textbf{43.7}$_{3.1}$ & -  & - \\
& \lmixin-\toxt & 86.9$_{1.1}$ & 85.5$_{0.3}$ & 11.2$_{1.7}$ & 97.6$_{0.3}$ & \textbf{71.4}$_{0.0}$ & 90.4$_{1.8}$ & 44.5$_{1.5}$ & -  & - \\
& \lmixin-AAE & -  & -  & \multicolumn{1}{c}{-}  & -  & -  & - & - & 92.3$_{0.1}$ & {16.1}$_{0.4}$ \\
\midrule[0.03em]
\multirow{5}{*}{\rotatebox{90}{{33\% train}}} 
& Random & 92.2$_{0.1}$ & 89.5$_{0.4}$ & 9.3$_{0.7}$ & \textbf{98.9}$_{0.0}$ & \textbf{83.3}$_{3.4}$ & 97.4$_{0.1}$ & 67.2$_{0.6}$ & 92.2$_{0.1}$ & 16.7$_{0.6}$\\
& AFLite & 91.9$_{0.1}$ & \textbf{90.2}$_{0.4}$ & 11.3$_{1.1}$ & 98.9$_{0.0}$ & 85.7$_{0.0}$ & 97.3$_{0.1}$ & 68.0$_{3.4}$ & 91.9$_{0.1}$ & 16.8$_{0.8}$\\
& DataMaps-Ambig. & 92.5$_{0.1}$ & 89.2$_{0.7}$ & 7.4$_{1.0}$ & 98.9$_{0.0}$ & 85.7$_{0.0}$ & \textbf{97.5}$_{0.0}$ & 64.4$_{1.4}$ & 92.5$_{0.1}$ & 16.0$_{0.4}$\\
& DataMaps-Hard & \textbf{92.6}$_{0.1}$ & 89.5$_{0.4}$ & {6.3}$_{0.9}$ & 98.8$_{0.0}$ & 85.7$_{0.0}$ & 97.4$_{0.0}$ & 62.0$_{1.1}$ & \textbf{92.6}$_{0.1}$ & \textbf{13.7}$_{0.2}$\\
& DataMaps-Easy & 91.9$_{0.2}$ & 86.8$_{0.6}$ & \textbf{5.9}$_{0.7}$ & 98.9$_{0.0}$ & \textbf{83.3}$_{3.4}$ & 97.2$_{0.1}$ & \textbf{60.3}$_{3.8}$ & 91.9$_{0.2}$ & 19.5$_{2.8}$\\
\midrule[0.03em]
\multicolumn{10}{l}{\textit{Ours (RoBERTa-base)}} \\
\midrule[0.03em]
& Vanilla & \bf 91.7$_{0.1}$ & \bf 90.1$_{0.3}$ & 8.4$_{0.4}$ & \bf 98.6$_{0.0}$ & 81.0$_{3.4}$ & 97.0$_{0.0}$ & 63.4$_{1.4}$ & \bf 95.9$_{0.2}$ & 16.9$_{1.0}$\\
& lexical removal & 90.9$_{0.0}$ & 86.0$_{0.7}$ & 18.3$_{1.5}$ & 98.1$_{0.1}$ & 78.6$_{0.0}$ & 96.4$_{0.0}$ & 61.7$_{0.2}$ & 95.1$_{0.1}$ & 18.7$_{0.6}$ \\
& InvRat (lexical) & 91.0$_{0.5}$ & 85.5$_{1.6}$ & \bf 3.4$_{0.6}$ & 97.5$_{1.0}$ & 76.2$_{3.4}$ & \bf 97.2$_{0.2}$ & 61.1$_{1.5}$ & 95.0$_{0.5}$ & 19.6$_{1.0}$\\
& InvRat (dialect) & 91.0$_{0.1}$ & 85.9$_{0.7}$ & \bf 3.4$_{0.5}$ & 97.6$_{0.5}$ & \bf 71.4$_{5.8}$ & 97.1$_{0.1}$ & \bf 57.9$_{2.2}$ & 93.1$_{1.0}$ & \bf 14.0$_{1.2}$\\

\bottomrule

\end{tabular}
\caption{\label{tab:results_lexical}
Evaluation of all debiasing methods on the \citet{founta2018large} test set. We show the mean and s.d. (subscript) of $F_1$ and FPR across 3 runs. 
The top two sections contain the scores reported in \citet{zhou2021challenges}. The bottom section contains scores of our methods.
When FPR is lower, the model is less biased by lexical associations for toxicity. We used RoBERTa-base, while RoBERTa-large is used in \citet{zhou2021challenges}. Thus, our Vanilla $F_1$ score is slightly lower than that of \citet{zhou2021challenges} by 0.5\%. 
}
\end{table*}

In the left four columns of Table~\ref{tab:results_lexical}, we show the $F_1$ scores and FPR in the entire dataset and in the \noi, \oi, and \oni attributes for measuring lexical bias. In addition to Vanilla, we include \emph{lexical removal}, a naive baseline that simply removes all words existing in \textsc{ToxTrig} before training and testing.

For our \InvRat (lexical/dialect) model, we can see a significant reduction in the FPR of \noi, \oi, and \oni over Vanilla (RoBERTa without debiasing). Our approach also yields consistent and usually more considerable bias reduction in all three attributes, compared to the ensemble and data filtering debiasing baselines discussed in \citet{zhou2021challenges}, where no approach improves in more than two attributes (e.g., \lmixin-\oni reduces bias in \oni but not the rest two; DataMaps-Easy improves in \noi and \oni but has similar FPR to Vanilla in \oi). The result suggests that \InvRat can effectively remove the spurious correlation between mentioning words in three lexical attributes and toxicity. Moreover, our \InvRat debiasing sacrifices little TLD performance\footnote{There is some degradation in \noi, which may result from some performance fluctuation in the small dataset and the labeling issues mentioned in \citet{zhou2021challenges}. We see the degradation as an opportunity for future dive deep rather than concerns. 
}, which can sometimes be a concern for debiasing (e.g., the overall performance of \lmixin). It is worth noting that the lexical removal baseline does not get as much bias reduction as our method, even inducing more bias in \noi. We surmise that the weak result arises from the limitation of \textsc{ToxTrig}, since a word bank cannot enumerate all biased words, and there are always other terms that can carry the bias to the model.

We summarize the debiasing results for the dialectal attribute in the rightmost column of Table~\ref{tab:results_lexical}. Compared with the Vanilla model, our method effectively reduces the FPR of AAE, suggesting the consistent benefit of \InvRat in debiasing dialect biases. Although the results from data relabeling \citep{zhou2021challenges} and some data filtering approaches are better than \InvRa, these approaches are complementary to \InvRa, and combining them presumably improves debiasing performance.

\subsection{Qualitative Study}

We demonstrate how \InvRat removes biases and keeps detectors focusing on genuine toxic clues by showing examples of generated rationales in Table~\ref{tab:qual}. 
Part (a) of Table~\ref{tab:qual} shows two utterances where both the baseline and our \InvRat debiasing predict the correct labels. We can see that when toxic terms appear in the sentence, the rationale generator will capture them. 
In part (b), we show three examples where the baseline model incorrectly predicts the sentences as toxic, presumably due to some biased but not toxic words (depend on the context) like \emph{\#sexlife, Shits, bullshit}. However, our rationale generator rules out these words and allows the TLD model to focus on main verbs in the sentences like \emph{keeps, blame, have}. In part (c), we show some examples that our \InvRat model fails to generate the true answer, while the baseline model can do it correctly. In these two examples, we observe that our rationale generator remove the offensive words, probably due to the small degree of toxicity, while the annotator marked them as toxic sentences.
Part (d) of Table~\ref{tab:qual} shows another common case that when the sentence can be easily classified as non-toxic, the rationale generator tends not to output any words, and the TLD model will output non-toxic label. It is probably caused by the non-stable predictive power of these non-toxic words (they are \emph{variant}), so the rationale generator choose to rule them out and keep rationale clean and invariant.

\newcommand{\tox}{\faExclamationTriangle}
\newcommand{\ton}{\faHandPeaceO}
\newcommand{\rat}[1]{\underline{\textbf{#1}}}

\begin{table*}[t]
\centering
\footnotesize
\begin{tabular}{c@{}p{10cm}ccc}
\toprule
& & Gold & Vanilla & Ours \\
\midrule
\multirow{2}{*}{(a) }& Oh my \rat{god} there's a \rat{f**king} STINKBUG and it's \rat{in my ASS} & \tox & \tox & \tox \\
\multirow{2}{*}{}&@user yes I hear that it's \rat{great} for a relationship to try and change your partner.. & \ton & \ton & \ton \\
\midrule
\multirow{3}{*}{(b) }&Other than \#kids, what \rat{keeps} you from the \#sexlife you want? & \ton & \tox & \ton \\
\multirow{3}{*}{}&Shits crazy but bet they'll \rat{blame}  us... wait for it & \ton & \tox & \ton \\ 
\multirow{3}{*}{}&@user @user You don't \rat{have} to pay for their bullshit read your rights read the law I don't pay fo… & \ton & \tox & \ton \\
\midrule
\multirow{2}{*}{(c) }&RT @user: my ex so ugly to me now like...i'll \rat{beat} that hoe ass & \tox & \tox & \ton \\
\multirow{2}{*}{}&@user \rat{Stop} that, it's not your \rat{fault} a scumbag decided to steal otems which were obviously meant for someone i… & \tox & \tox & \ton \\
\midrule
\multirow{1}{*}{(d) }& A shark washed up in the street after a cyclone in Australia & \ton & \ton & \ton \\
\bottomrule
\end{tabular}
\caption{\label{tab:qual}
Examples from the test set with the predictions from vanilla and our models.
\tox~denotes toxic labels, and \ton~denotes non-toxic labels.
The underlined words are selected as the rationale by our ratinoale generator.
}
\end{table*}

\section{Conclusion}

In this paper, we propose to use \InvRat to reduce the biases in the TLD models effectively. By separately using lexical and dialectal attributes as the environments in \InvRat framework, the rationale generator can learn to generate genuine linguistic clues and rule out spurious correlations.
Experimental results show that our method can better mitigate both lexical and dialectal biases without sacrificing much overall accuracy.
Furthermore, our method does not rely on complicated data filtering or relabeling process, so it can be applied to new datasets without much effort, showing the potential of being applied to practical scenarios.

\bibliographystyle{acl_natbib}
\bibliography{acl2021}

\clearpage
\newpage

\appendix
\section{Bias attributes}
\label{app:attributes}

We follow \citet{zhou2021challenges} to define four attributes (\noi, \oi, \oni, and \aae) that are often falsy related to toxic language. \noi is mention of minoritized identities (e.g., \emph{gay, female, Muslim}); \oi mentions offensive words about minorities (e.g., \emph{queer, n*gga}); \oni is mention of swear words (e.g., \emph{f*ck, sh*t}). \noi should not be correlated with toxic language but is often found in hateful speech towards minorities~\citep{dixon2018measuring}. Although \oi and \oni can be toxic sometimes, they are used to simply convey closeness or emphasize the emotion in specific contexts~\citep{dynel2012swearing}. \aae contains dialectal markers that are commonly used among African Americans. Even though \aae simply signals a cultural identity in the US \citep{green2002african}, \aae markers are often falsy related to toxicity and cause content by Black authors to mean suppressed more often than non-Black authors \citep{sap2019risk}.

\section{Training Details}
\label{app:details}

We use a single NVIDIA TESLA V100 (32G) for each experiment. The average runtime of experiments for \emph{Vanilla} model in Table~\ref{tab:results_lexical} are 2 hours. The \InvRat model in Table~\ref{tab:results_lexical} need about 9 hours for a single experiment.

The main hyperparameters are listed in Table~\ref{tab:hyper}. More details can be found in our released code.
We did not conduct hyperparameter search, but follow all settings in the official implementation of \citet{zhou2021challenges} \footnote{\url{https://github.com/XuhuiZhou/Toxic_Debias}}.
One difference is that because \InvRat framework needs three RoBERTa models to run at the same time, we choose to use RoBERTa-base, while \citet{zhou2021challenges} uses RoBERTa-large. As a result, our $F_1$ score for the Vanilla model is about 0.5 less than the score in \citet{zhou2021challenges}.

\begin{table}[t!]
\centering
\small
\setlength\tabcolsep{3pt}
\begin{tabular}{cc}
\toprule
\bf hyperparameter & \bf value \\
\midrule
optimizer & AdamW \\
adam epsilon & $1.0 \times 10^{-8}$ \\
learning rate & $1.0 \times 10^{-5}$ \\
training epochs & 10 \\
batch size & 8 \\
max gradient norm & 1.0 \\
weight decay & 0.0 \\
sparsity percentage ($\alpha$) & 0.2 \\
sparsity lambda ($\lambda_1$) & 1.0 \\
continuity lambda ($\lambda_2$) & 5.0 \\
diff lambda ($\lambda_{\text{diff}}$) & 10.0 \\
\bottomrule
\end{tabular}
\caption{The main hyperparameters in the experiment. Sparsity percentage is the value of $\alpha$ in $\mathcal{L}_{reg}$ mentioned in equation~\ref{eq:sparsity}; sparsity lambda and continuity lambda are $\lambda_1$ and $\lambda_2$ in equation~\ref{eq:sparsity}; diff lambda is $\lambda_{\text{diff}}$ in equation~\ref{eq:all}.}
\label{tab:hyper}
\end{table}

\end{document}